\documentclass{article}
\usepackage{spconf,amsmath,graphicx}
\usepackage{booktabs}
\usepackage[utf8x]{inputenc} 
\usepackage{multirow}
\usepackage{multicol}
\usepackage{bm}

\title{ESCL: Equivariant Self-Contrastive Learning for Sentence Representations}
\name{Jie Liu$^{1*}$\qquad Yixuan Liu$^{1,2*}$\qquad Xue Han$^1$\qquad Chao Deng$^1$\qquad Junlan Feng$^{1*}$\thanks{$^*$ The first two authors contribute equally.}
\thanks{$^*$ Junlan Feng is the corresponding author.}
}
\address{$^1$JIUTIAN Team, China Mobile Research \\ $^2$Beijing University of Posts and Telecommunications}
\begin{document}
%
\maketitle
\begin{abstract}
Previous contrastive learning methods for sentence representations often focus on insensitive transformations to produce positive pairs, but neglect the role of sensitive transformations that are harmful to semantic representations. Therefore, we propose an Equivariant Self-Contrastive Learning (ESCL) method to make full use of sensitive transformations, which encourages the learned  representations to be sensitive to certain types of transformations with an additional equivariant learning task. Meanwhile, in order to improve practicability and generality, ESCL simplifies the implementations of traditional equivariant contrastive methods to share model parameters from the perspective of multi-task learning. We evaluate our ESCL on semantic textual similarity tasks. The proposed method achieves better results while using fewer learning parameters compared to previous methods.
\end{abstract}
\section{Introduction}
\label{sec:intro}
Sentence representation is a fundamental task in the field of natural language processing, which has been well studied in previous literatures \cite{sent_emb_2017,sent_bert_2019,scd,sncse}. In practice, sentence embeddings are widely used in numerous downstream tasks, such as text summarization \cite{text_summarization_2017}, machine translation \cite{machine_translation_2017} and recommendations \cite{metricbert}. Recently, some studies found that fine-tuning Pre-trained Language Models (PLMs) \cite{bert} with contrastive learning is helpful to learn sentence embeddings \cite{self-guided,virt,ease,virtual}. Typically, contrastive learning methods construct positive pairs through data augmentations while treating other unrelated samples as negative instances, and then improve the representation space of PLMs based on InfoNCE loss \cite{simclr}. Existing contrastive learning methods treat data augmentation modules as insensitive transformations that cannot affect the semantic representation (e.g., image blurring, low-dropout-based augmentation), but ignore the role of sensitive transformations that are harmful to semantic representation \cite{equivarint_contrstive_learning} (e.g., image rotations and word deletions). That is, sentence representations learned through fine-tuning PLMs with a contrastive learning strategy should be sensitive to certain types of transformations.

Based on the idea of contrastive learning, SimCSE \cite{simcse} simplifies its implementation by only using standard dropout as an implicit data augmentation.
In this work, inspired by SimCSE and equivariant self-supervised learning methods \cite{equivarint_contrstive_learning, diffcse}, we propose an Equivariant Self-Contrastive Learning (ESCL) method that relies only on dropout-based data augmentation to improve the expressiveness of sentence  representations. Following SimCSE, the proposed ESCL uses the dropout-based data augementation with low dropout rate as insensitive transformation to bulid an invariant task (similar to the main task in multi-task learning \cite{multi_task_learning}). In the framework of equivariant self-supervised learning \cite{equivarint_contrstive_learning}, we construct the equivariant task (similar to the auxiliary task) using high dropout rate and the proposed Relative Difference (RD) loss. From the view of multi-task learning, 
we analyze equivariant self-supervised learning in the hope of making it more practical and providing researchers with a new perspective.

\section{Related Work}
\label{sec:format}
Most of the contextualized neural embedding methods are based on PLMs and show great promise. 
However, their sentence representations cannot achieve satisfactory performance on downstream tasks. 

Some recent studies use a contrastive learning strategy to fine-tune PLMs to get better sentence embeddings. DeCLUTR \cite{declutr} adopted a span sampling method in the same document to get anchor spans and positive spans.
Self-guided contrastive framework \cite{self-guided} cloned BERT into two copies to get multiple views of the same sample. ConSERT \cite{consert} verified the effectiveness of multiple text augmentation strategies.
SimCSE \cite{simcse} used only standard dropout in PLMs twice as implicit data augmentations.
SNCSE \cite{sncse} proposed the soft negative samples 
and a bidirectional margin loss to distinguish and decouple textual similarity and semantic similarity. 

\begin{figure}[htb]
  \centering
  \centerline{\includegraphics[width=8.5cm]{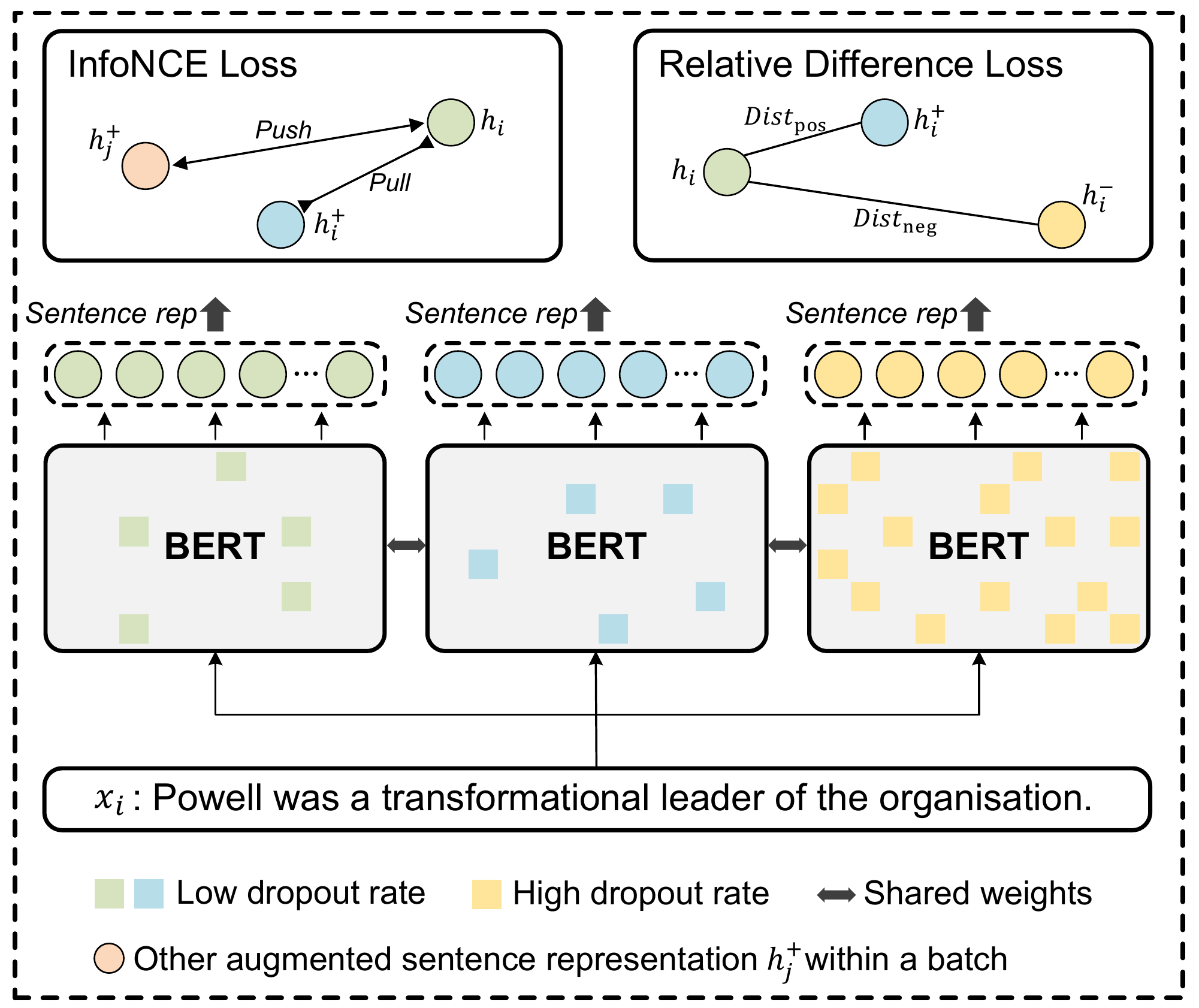}}

\caption{Schematic illustration of the proposed method.}
\label{fig:fig_1}
\end{figure}

More recently, to make full use of the previously ignored sensitive transformations,
E-SSL \cite{equivarint_contrstive_learning} added an additional task to contrastive learning framework to make the learned embeddings more expressive in the field of computer vision. Subsequently, DiffCSE \cite{diffcse} applied this idea 
to sentence representations. However, DiffCSE employs an additional generator to produce augmented samples and a discriminator to build the equivariant task, which not only makes the computation more expensive, but also leads to more complex model structures and more training parameters. Compared to E-SSL and DiffCSE, our ESCL is more efficient since it does not need additional data augmentation modules and encoders, and only uses the dropout-based data augmentations to construct invariant and equivariant tasks.

\section{Methodology}
\label{sec:pagestyle}

\subsection{General Contrastive Learning Framework}
In a typical contrastive learning method \cite{simclr}, the training objective is designed to obtain effective representation by pulling similar samples closer while pushing the unrelated samples apart.

SimCSE assumes a minibatch of $\mathit{N}$ samples $ \mathcal{D} = \left \{ x_{i}  \right \} _{i=1}^{N} $, where $x_{i}$ denotes the $\mathit{i}$-th input sentence. SimCSE passes $x_{i}$ to BERT  with the same low dropout rate twice to get two sentence embeddings $h_{i}$ and $h_{i}^{+}$, which is equivalent to using two different sub-encoders from original BERT. That is, unsupervised SimCSE is an implicit parameter-shared dual-encoder framework. As shown in Fig. \ref{fig:fig_1}, the embeddings of positive pair for the given sentence $x_{i}$ can be obtained by:
\begin{equation}\label{equ1}
h_{i}=f_{\theta }  (x_{i},r_{\rm low},m_{i}),\enspace h_{i}^{+} =f_{\theta }  (x_{i},r_{\rm low},m_{i}^{+})
\end{equation}
where $\theta$ are the training parameters of encoder $f$, $m_{i}$ and $m_{i}^{+}$ denote different dropout masks for the low dropout rate $r_{\rm low}$. The InfoNCE loss for input sentence $x_{i}$ in a mini-batch $\mathcal{D}$ can be formulated as follows:
\begin{equation}\label{equ2}
\mathcal{L}_{\rm InfoNCE} =- \log_{}{ \frac{e^{{\rm sim}\left ( h_{i}, h_{i}^{+}  \right )/\tau  } }{ {\textstyle \sum_{j= 1}^{N} }{\textstyle e^{{\rm sim}(h_{i}, h_{j}^{+})/\tau }} }} 
\end{equation}
where $\tau$ is a temperature hyperparameter and $\rm{sim(\cdot,\cdot)}$ is the  cosine similarity.
The training objective treats other $N-1$ augmented samples within a minibatch as negative samples and aims to distinguish positive samples from negative ones, even if the difference of the two is small. In other words, the hard negative samples play an important role in InfoNCE loss.

\subsection{Equivariant Self-Contrastive Learning}
\label{subsec:equivariant}
More recently, E-SSL \cite{equivarint_contrstive_learning} proposed a general equivariant self-supervised learning framework, which discussed and verified the importance of previously neglected sensitive transformations for learning sentence representation in the field of computer vision. Let $T_{g}$ denote the transformation from a group $G$, $T_{g}^{\prime}$ denotes an induced group transformation, $f$ is the encoder to get representations and $x$ is an input sample. The property of equivariance can be described as:
\begin{equation}\label{equ4}
f(T_{g}(x))=T_{g}^{\prime}(f(x)) 
\end{equation}
We can construct a training objective to make $T_{g}^{\prime}$ not the identity for some types of transformations (e.g., image rotations), while it can keep the identity for some other transformations (e.g., image blurring).

In equivariant self-supervised learning, we usually need to construct an equivariant task. E-SSL directly adopts data augmentation to get the augmented samples, which have different semantics from the original samples. DiffCSE uses an additional generator to produce augmented sentences and an additional discriminator encoder with new training parameters to build the equivariant task. 
In contrast to these above methods, for the sake of efficiency, we use the above encoder $f$ with high dropout rate to accomplish sensitive transformation to get the embedding $h_{i}^{-}$. As shown in Fig. \ref{fig:fig_1}, $h_{i}^{-}$ can be obtained by:
\begin{equation}\label{equ5}
h_{i}^{-}=f_{\theta }  (x_{i},r_{\rm high},m_{i}^{-})
\end{equation}
where $r_{\rm high}$ is a high dropout rate and $m_{i}^{-}$ denotes its dropout mask. That is to say, we construct the equivariant task using only dropout-based data augmentation with a high dropout rate. With no need for  the additional data augmentation module \cite{equivarint_contrstive_learning, diffcse} and discriminator \cite{diffcse} to construct the equivariant task, our ESCL can simplify the model structure and reduce the scale of training parameters.

Based on the property of equivariant self-supervised learning and inspired by SNCSE \cite{sncse}, we design a Relative Difference (RD) loss for sensitive transformations denoted by $\mathcal{L}_{\rm RD}$, which aims to learn the relative difference between positive and negative samples. The RD loss  is defined as:
\begin{equation}\label{equ6}
\mathcal{L}_{\rm RD} = {\textstyle \sum_{h_{i}^{\prime}\in \left \{  h_{i}, h_{i}^{+}\right \} }^{}} e^{{\rm sim}(h_{i}^{\prime}, h_{i}^{-})-{\rm sim}(h_{i}, h_{i}^{+})}
\end{equation}
Relative difference loss function $\mathcal{L}_{\rm RD}$ encourages the cosine distance ${Dist}_{\rm neg}$ between negative pair ($h_{i}^{\prime}$ and $h_{i}^{-}$) to be much larger than the cosine distance ${Dist}_{\rm pos}$ between positive pair ($h_{i}$ and $h_{i}^{+}$). 
The training objective design is based on the property of equivariant contrastive learning, which helps the learned sentence embeddings be sensitive to certain types of transformations that are harmful to semantic representation.

As mentioned above, we can get the final loss function $\mathcal{L}_{\rm ESCL}$ which consists of two training objectives:
\begin{equation}\label{equ7}
\mathcal{L}_{\rm ESCL} = \mathcal{L}_{\rm InfoNCE} + \lambda \cdot \mathcal{L}_{\rm RD}
\end{equation}
where $\lambda$ is a hyperparameter to control the trade-off between these two loss functions. All the training procedures of our ESCL are described as above and illustrated in Fig. \ref{fig:fig_1}.

In the inference stage, we discard the equivariant task and use only the encoder $f$ to produce sentence embeddings.

Another advantage is that the structure of ESCL is similar to the framework of hard parameter sharing for multi-task learning in deep neural networks \cite{multi_task_learning}, which shares the training parameters for different tasks that can promote each other during training. 
Although the invariant task and equivariant task do not exactly meet the requirements of multi-task learning, the similarity in the framework makes many of the studies of multi-task learning useful for equivariant contrastive learning. We hope that, from the view of multi-task learning, we can provide a new research perspective for equivariant contrastive learning.

$\textbf{Why does the relative difference loss work?}$ To further understand the role of $\mathcal{L}_{\rm RD}$, we analyze and compare InfoNCE loss and RD loss. Firstly, InfoNCE loss in Eq. \ref{equ2} can be formulated in another way:
\begin{equation}\label{equ8}
\mathcal{L}_{\rm InfoNCE} =\log({1+\frac{{\textstyle \sum_{j=1, j\ne i}^{N} }{\textstyle e^{{\rm sim}(h_{i}, h_{j}^{+})/\tau }}}{e^{{\rm sim}\left ( h_{i}, h_{i}^{+}  \right )/\tau  }}})
\end{equation}
It is clear that cosine distance ${Dist}_{\rm pos}$ should be smaller, while ${Dist}_{\rm neg}^{\prime}$ between $h_{i}$ and $h_{j}^{+}$ should be larger. However, InfoNCE loss may cause some problems: (i). The negative samples come from the same batch, so there may be some false negative samples, which will affect the effect of InfoNCE loss. (ii). There is no explicit comparison between ${Dist}_{\rm pos}$ and ${Dist}_{\rm neg}^{\prime}$. Compared to InfoNCE loss, RD loss in Eq. \ref{equ6} explicitly encourages ${Dist}_{\rm neg}^{\prime}$ to be greater than ${Dist}_{\rm pos}$, and the embeddings of negative samples come from BERT with high dropout to ensure quality. Therefore, RD loss can enable BERT to make full use of the sensitive  transformations to get better sentence embeddings.

\section{Experiments and Analysis}
\label{sec:typestyle}

\subsection{Experimental Setup}
In our experiment, we implement our ESCL based on the PyTorch implementations of SimCSE \cite{simcse} and DiffCSE \cite{diffcse}. Following the setting of DiffCSE, we use BERT(uncased) \cite{bert} to initialize the sentence encoder $ f $ at the training stage. Unless otherwise mentioned, the rest of the hyperparameters in our ESCL are the same as in DiffCSE \cite{diffcse}. We use Spearman’s correlation $\rho$ to measure the performance of the learned sentence embeddings, which is a non-parametric measure of rank correlation and can be formulated as:
\begin{equation}\label{equ9}
\rho (\mu ,\nu )=\frac{\sum_{k=1}^{n} (\mu _{k}- \bar{\mu} )(\nu _{k}- \bar{\nu} )}{\sqrt{\sum_{k=1}^{n} (\mu _{k}- \bar{\mu} )^{2}\sum_{k=1}^{n} (\nu _{k}- \bar{\nu})^{2} } }
\end{equation}
where $\mu$ and $\nu$ are a set of variables, $n$ is the sample size, $\mu _{k}$ and $\nu _{k}$ denote the $k$-th variable, $\bar{\mu}$ and $\bar{\nu}$ denote the mean value.

For the additional hyperparameters in our ESCL, we set $ r_{\rm low}$ as $0.1$, $r_{\rm high} $ as $0.45$ and $ \lambda $ as $ 2.5e-3 $. We will compare the results of using different $ r_{\rm high} $ for the equivariant learning task in Sec.  \ref{subsec:ablation}.
In subsequent sections, we report the performance of our ESCL over $ 10 $ different random seeds to reduce statistical errors.

\subsection{The Datasets}
We use the SentEval \cite{senteval} toolkit to evaluate ESCL on 7 semantic textual similarity (STS) tasks, which include STS 2012-2016 \cite{sts-2016}, STS Benchmark \cite{sts-b} and SICK-Relatedness \cite{sick-r}. It is worth mentioning that no STS training datasets are used at the training stage and all the experiments on STS are fully unsupervised, which means all the embeddings are fixed once they are trained. We choose to follow the way of using development data of Sentence-BERT \cite{sent_bert_2019} in our evaluation.
SimCSE and DiffCSE also use the same strategy in evaluation.

\subsection{Main Results and Analysis}  
{\bf{Baselines.}}
We compare our ESCL to previous state-of-the-art methods on STS tasks including averaged GloVe \cite{glove} embeddings, averaged first and last layer BERT \cite{bert} embeddings, SimCSE \cite{simcse}, DiffCSE \cite{diffcse} and the post-processing method BERT-flow \cite{bert-flow}. Tab. \ref{table:table_1} shows all the related results on 7 STS tasks for different methods based on BERT$_{\texttt{\scriptsize base}}$ \footnote{Additionally, we repeat all the experiments based on RoBERTa$_{\texttt{\scriptsize base}}$, which also proved the effectiveness of ESCL.}.

Firstly, compared to the axiomatic method GloVe, our ESCL achieves a significant performance improvement on all STS datasets, which fully demonstrates the effectiveness of the contextualized neural embedding methods based on PLMs with the contrast learning strategy. 

\begin{table*}[t]
    \centering
    \renewcommand\arraystretch{1.25}
    \begin{tabular}{l   c   c   c   c   c   c   c   c}
        \toprule
        \textbf{Model}  &\textbf{STS12} &\textbf{STS13} &\textbf{STS14} &\textbf{STS15} &\textbf{STS16} &\textbf{STS-B} &\textbf{SICK-R}    &\textbf{Avg.}
        \\ \hline
        GloVe embeddings (avg.)$^\dagger$\quad\quad\quad\quad\quad\quad  & 55.14 & 70.66 & 59.73 & 68.25 & 63.66 & 58.02 & 53.76    & 61.32 \\
        BERT$_{\texttt{\scriptsize base}}$  (first-last avg.)$^\dagger$   & 39.70 & 59.38 & 49.67 & 66.03 & 66.19 & 53.87 & 62.06    & 56.70 \\ 
        BERT$_{\texttt{\scriptsize base}}$-flow$^\dagger$   & 58.40 & 67.10 & 60.85 & 75.16 & 71.22 & 68.66 & 64.47    & 66.55 \\ 
        \hline
        \texttt{$\ast$} SimCSE$_{\texttt{\scriptsize cls}}$ {\footnotesize(reproduce)}  & \textbf{68.21} & 81.32 & 73.72 & 80.25 & 76.03 & 75.54 & 71.06    & 75.16 \\
        \texttt{$\ast$} DiffCSE$_{\texttt{\scriptsize cls}}$ {\footnotesize(reproduce)}     & 66.42 & 81.60 & 73.46 & \textbf{82.29} & 78.00 & 77.22 & 70.29    & 75.61 \\
        ESCL$_{\texttt{\scriptsize cls}}$   & 66.67 & \textbf{82.66} & \textbf{74.03} & 82.24 & \textbf{79.78} & \textbf{79.49} & \textbf{72.46}    & \textbf{77.19} \\
        \hline
        \texttt{$\ast$} SimCSE$_{\texttt{\scriptsize cls-before-pooler}}$ {\footnotesize(reproduce)}   & 68.06 & 81.56 & 73.95 & 80.84 & 76.56 & 75.79 & 71.43    & 75.46 \\
        \texttt{$\ast$} DiffCSE$_{\texttt{\scriptsize cls-before-pooler}}$ {\footnotesize(reproduce)}   & 67.21 & 81.84 & 74.06 & 82.62 & 78.97 & 77.57 & 70.82    & 76.16 \\
        ESCL$_{\texttt{\scriptsize cls-before-pooler}}$ & \textbf{70.06} & \textbf{82.64} & \textbf{74.14} & \textbf{82.67} & \textbf{80.14} & \textbf{80.14} & \textbf{72.44}    & \textbf{77.46} \\
        \bottomrule
    \end{tabular}
    \caption{The performance of sentence representations on semantic textual similarity (STS) test sets (Spearman’s correlation) for different methods based on BERT$_{\texttt{\scriptsize base}}$. $ \dagger $ means the result comes from DiffCSE, \texttt{$\ast$} means the reproduced results with default setup based on the original implementations of SimCSE\protect\footnotemark[1] and DiffCSE\protect\footnotemark[2].}
\label{table:table_1}
\end{table*}

\begin{table}[t]
    \centering
    \renewcommand\arraystretch{1.25}
    \tabcolsep=0.15cm
    \begin{tabular}{l   c   c   c   c   c}
        \toprule
        \multirow{2}{*}{\textbf{Setup}}  &\multicolumn{4}{c}{\bm{$r_{\rm high}$}}  &\textbf{Equivariant Loss} \\
        ~ &\textbf{$0.35$} &\textbf{$0.40$} &\textbf{$0.45$} &\textbf{$0.50$}  &\text{CosSim loss} \\
        \hline
        \textbf{STS-B}  & 82.01 & 83.74 & \textbf{83.94} & 83.45 & 80.29 \\
        
        \bottomrule
    \end{tabular}
    \caption{Development set results of STS-B with different dropout rates and loss in equivariant task.}
\label{table:table_2}
\end{table}

Compared to the contextualized neural embedding methods based on PLMs (BERT, BERT-flow and SimCSE), our ESCL method still achieves consistent performance gains. As mentioned above, the original BERT is not suitable for directly getting sentence embeddings. BERT-flow is a post-processing method which  directly adjusts the anisotropic distribution of sentence embeddings through normalizing flows, limited by the flow-based model in PLMs, resulting in a relatively small performance improvement. Specifically, although SimCSE is also a method for fine-tuning BERT based on a contrastive learning strategy, ESCL outperforms it on STS tasks by about $2\%$ on Spearman’s correlation.

\footnotetext[1]{https://github.com/princeton-nlp/SimCSE}
\footnotetext[2]{https://github.com/voidism/DiffCSE}

Finally, the most important comparison of experimental results is between DiffCSE and our ESCL. DiffCSE is an equivariant contrast learning method, that uses an additional generator to produce augmented samples and a discriminator to construct the equivariant task for sensitive transformations. Our ESCL$_{\texttt{\scriptsize cls-before-pooler}}$ can also improve upon DiffCSE$_{\texttt{\scriptsize cls-before-pooler}}$ from $76.16\%$ to $77.46\%$. Such experimental results fully validate our analysis of building the equivariant task in Sec. \ref{subsec:equivariant}.

\subsection{Ablation Studies}
\label{subsec:ablation}
In this section, we present a series of ablation experiments to support the reasonability of the design of our ESCL in Tab. \ref{table:table_2}. The following variants are considered: (i). $r_{\rm high}$ in Eq. \ref{equ5} for augmented sentence embeddings. (ii). The loss function of equivariant tasks. 

For the high dropout rate $r_{\rm high}$, which is a hyperparameter that affects the quality of the embedding $h_{i}^{-}$ to build the equivariant task. To further understand the role of $r_{\rm high}$ in Eq. \ref{equ5}, we try the different values in Tab. \ref{table:table_2} and observe that the augmented embedding $h_{i}^{-}$ from BERT with a high dropout rate plays an important role in the equivariant task, and the way of building augmented embeddings for sensitive transformations is effective. Therefore, we set $r_{\rm high}$ as $0.45$ in all experiments for our ESCL.

Then, we replace the RD loss with a simple Cosine Similarity (CosSim) loss $ {\textstyle \sum_{h_{i}^{\prime}\in \left \{  h_{i}, h_{i}^{+}\right \} }^{}}  e^{sim(h_{i}^{\prime}, h_{i}^{-})}$ to verify the role of RD loss in the equivariant task. The CosSim loss aims to make the cosine distance ${Dist}_{\rm neg}$ between the negative pair ($h_{i}^{\prime}$ and $h_{i}^{-}$) larger, but cannot get the relative differences between ${Dist}_{\rm pos}$ and ${Dist}_{\rm neg}$. As shown in Tab. \ref{table:table_2}, even though CosSim loss takes the same $h_{i}^{-}$ as augmented sentence embeddings, the performance degrades by $3.65\%$ on the development set of STS-B. This comparative experiment shows that the learned sentence representations by RD loss are sensitive to the difference between the original sample and augmented sample,  and the relative difference between ${Dist}_{\rm pos}$ and ${Dist}_{\rm neg}$ is conducive to improving the sentence representation of PLMs.

\section{Discussion and Conclusions}
\label{sec:majhead}
We introduce ESCL, an equivariant self-contrastive learning method that improves the sentence representations of BERT, which relies only on standard dropout-based augmentations.
Firstly, different dropout rates are used to build invariant and equivariant tasks.
Subsequently, the relative difference loss for the equivariant task is proposed to jointly optimize sentence representations.
Finally, we provide researchers with a new multi-task learning perspective to analyze and study equivariant contrastive learning.
We believe that our ESCL can provide a new framework to implement equivariant self-supervised learning to get better sentence embeddings.

\vfill\pagebreak

\bibliographystyle{IEEEbib}
\bibliography{strings,refs}

\end{document}